\documentclass[preprint]{elsarticle}
\usepackage{CJKutf8} 
\usepackage{times}  
\usepackage{helvet} 
\usepackage{courier}  
\usepackage{graphicx} 
\usepackage{caption} 

\usepackage{subfigure}
\usepackage{hyperref}
\usepackage[utf8]{inputenc}
\usepackage[english]{babel}

\usepackage{epsfig}
\usepackage{epstopdf}
\usepackage{amssymb}
\usepackage{amsmath}
\usepackage{multirow}
\usepackage{makecell}
\usepackage{tikz}
\usepackage{algorithm}
\usepackage{booktabs}
\usepackage{algorithmic}
\usepackage{array}
\usepackage{longtable}
\usepackage{stfloats}
\usepackage{supertabular}
\usepackage{lineno} 

\usepackage{color}

\newcommand{\jinkb}[1]{{\color{black} #1 \color{black}}}
\newcommand{\hankz}[1]{\color{black} #1 \color{black}}

\newcommand{\ignore}[1]{{}}

\usepackage{bibentry}

\newcommand{\zhmap}[1]{{\color{black} #1}}
\newcommand{\zhma}[1]{{\color{black} #1}}

\def\ours{\tt SORL}
\def\oursh{\tt SORL-HTN}

\begin{document}


\title{Creativity of AI: Hierarchical Planning Model Learning for Facilitating Deep Reinforcement Learning}
\author[mymainaddress]{Hankz Hankui Zhuo\corref{mycorrespondingauthor}}
\cortext[mycorrespondingauthor]{Corresponding author}
\ead{zhuohank@mail.sysu.edu.cn}

\author[mymainaddress]{Shuting Deng}
\ead{dengsht6@mail2.sysu.edu.cn}

\author[mymainaddress]{Mu Jin}
\ead{jinm6@mail2.sysu.edu.cn}

\author[mymainaddress]{Zhihao Ma}
\ead{mazhh7@mail2.sysu.edu.cn}

\author[mymainaddress]{Kebing Jin}
\ead{jinkb@mail2.sysu.edu.cn}

\author[mysecondaryaddress]{Chen Chen}
\ead{chenchen9@huawei.com}

\author[mymainaddress]{Chao Yu}
\ead{yuchao3@mail.sysu.edu.cn}

\address[mymainaddress]{School of Computer Science and Engineering, Sun Yat-sen University, Guangzhou, China}
\address[mysecondaryaddress]{Huawei Noah’s Ark Lab}

\begin{frontmatter}
\begin{abstract}
Despite of achieving great success in real-world applications, Deep Reinforcement Learning (DRL) is still suffering from three critical issues, i.e., data efficiency, lack of the interpretability and transferability. Recent research shows that embedding symbolic knowledge into DRL is promising in addressing those challenges. Inspired by this, we introduce a novel deep reinforcement learning framework with symbolic options. Our framework features a loop training procedure, which enables guiding the improvement of policy by planning with planning models (including action models and hierarchical task network models) and symbolic options learned from interactive trajectories automatically. The learned symbolic options alleviate the dense requirement of expert domain knowledge and provide inherent interpretability of policies. Moreover, the transferability and data efficiency can be further improved by planning with the symbolic planning models. To validate the effectiveness of our framework, we conduct experiments on two domains, Montezuma's Revenge and Office World, respectively. The results demonstrate the comparable performance, improved data efficiency, interpretability and transferability.
\end{abstract}
\end{frontmatter}

\section{Introduction}
Deep Reinforcement Learning (DRL) has achieved tremendous success in complex and high dimensional environments such as Go \cite{DBLP:journals/nature/SilverHMGSDSAPL16,silver2017mastering} and Atari Games \cite{mnih2015human}. It interacts with environments and improves its policy with the collected experience, by maximizing the long term reward. Recent criticism on DRL mostly focuses on the lack of transferability, interpretability, and data efficiency. 
The policy learnt from an environment often fails in another unseen environment. Due to the use of black-box neural networks for function approximation, the intrinsic lack of interpretability issue naturally raises in DRL, which disables the agent to explain its actions in a human-understandable way and earn people's trust in critical areas such as autonomous driving \cite{autonomousdriving} and chemical engineering \cite{zhou2017optimizing}. Besides, DRL often requires a large amount of data to learn a satisfying policy in complex environments. The process of collecting experiences for learning the policy is time-consuming and the sample efficiency is low. 


To alleviate those issues, 
researchers have investigated the combination of HRL and symbolic planning to improve transferability, interpretability, and data efficiency \cite{DBLP:conf/icml/Ryan02,DBLP:journals/ai/LeonettiIS16,DBLP:conf/ijcai/YangLLG18,DBLP:conf/aaai/LyuYLG19,DBLP:conf/aips/IllanesYIM20,sarathy2020spotter,lee2021ai}. In those approaches, the original MDP is divided into two levels. The higher level utilizes a symbolic planner with a given planning model to generate plans for selecting \emph{options}, while the lower-level interacts with the environment to accomplish the selected options. This two-level structure helps alleviating the sparse reward issue, and improves sampling efficiency with the help of generated plans.
In those work, however, they require the planning models have been provided by domain experts. 
In many real-world applications, however, it is often difficult to create planning models by hand \cite{DBLP:journals/ai/YangWJ07}, especially when the environment is complicated. A more realistic idea is to automatically learn planning models from training data \cite{DBLP:journals/ai/ZhuoK17,DBLP:DBLP:journals/ai/YangWJ07,DBLP:conf/ijcai/NgP19,DBLP:conf/aips/0004ATRI16,DBLP:conf/icml/JamesR020} and exploit the learnt planning models to generate plans for guiding the exploration of options. 
Although there is indeed an approach \cite{sarathy2020spotter} proposed to learn planning models automatically, they still need to manually define a major part of the models in advance. Besides, the planning goal in this approach is kept unchanged while it is dynamically adapted to maximize the external reward in our framework.

\begin{figure}[!ht]
    \centering
    \includegraphics[width=0.9\textwidth]{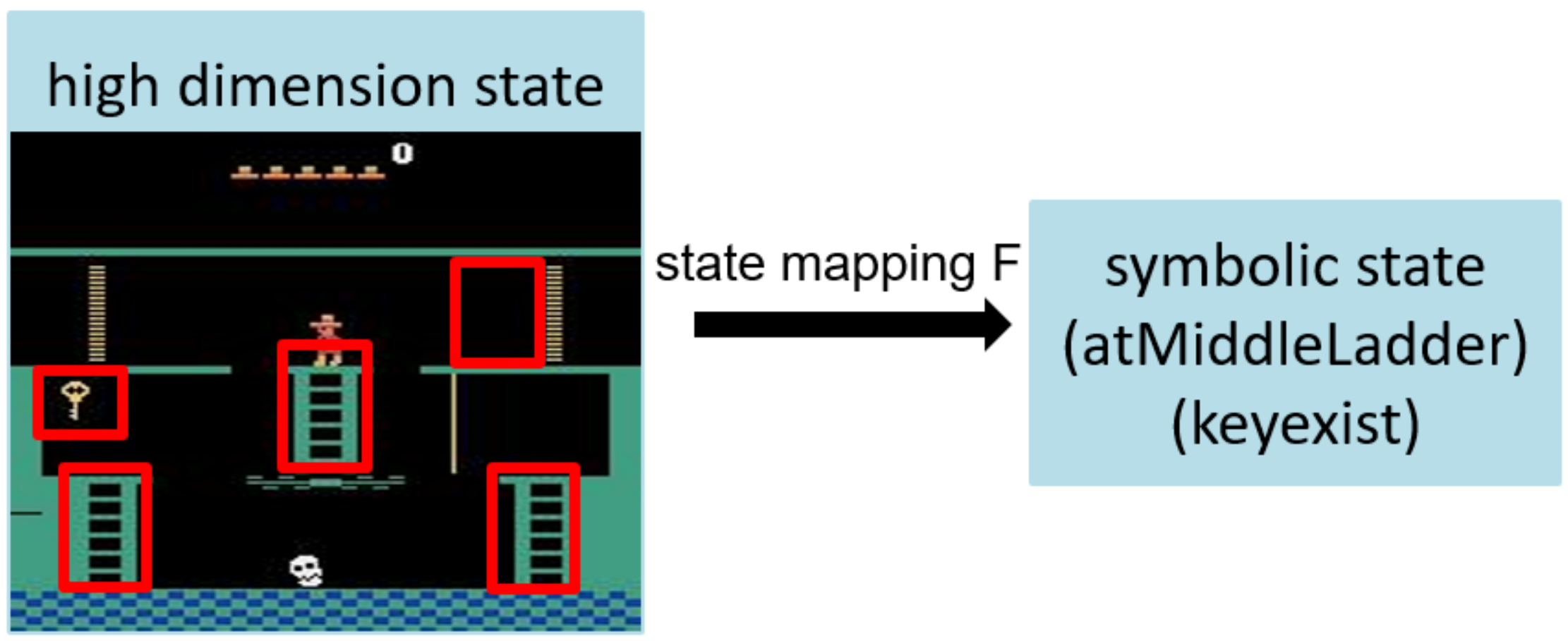}
    \caption{State Mapping Function in Montezuma's Revenge}
    \label{fig:statemap}
\end{figure}
\begin{figure}[!ht]
    \centering
    \includegraphics[width=8cm]{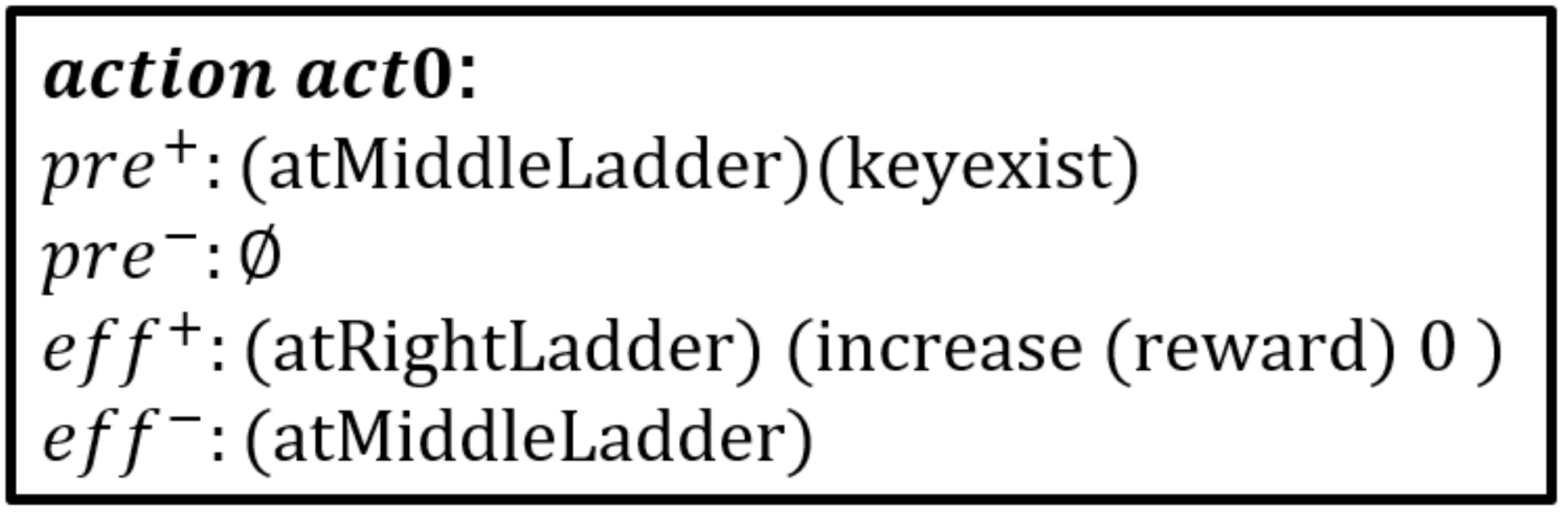}
    \caption{An Action model in Montezuma's Revenge}
    \label{fig:actionmodel}
\end{figure}

In our previous work \cite{DBLP:conf/aaai/JinMJZCY22}, we propose a novel framework, namely {\ours}, which stands for \textbf{S}ymbolic \textbf{O}ptions for \textbf{R}einforcement \textbf{L}earning, to learn action models to help the exploration of actions in reinforcement learning. We assume
that there exists a function $F$, mapping high dimensional states to symbolic states and enabling us to learn symbolic action models and options. As shown in Figure \ref{fig:statemap}, we extract the position of the man in red and the key from the high-dimensional state to obtain the corresponding symbolic state. 
When the agent walks from the middle ladder to the right ladder,
the key still exists and the environment does not give any feedback (e.g., zero reward). This can be seen as a symbolic transition and we can generate the corresponding action model as shown in Figure \ref{fig:actionmodel}. Then, we use a planner with the learned action models and a planning goal as input to generate a plan and use it to instruct the learning of the agent.

Based on the assumption, {\ours} features a two-level structure, of which the higher level is a symbolic planner and a meta-controller, and the lower level is an RL agent interacting with the environment. 
The higher level utilizes the collected trajectories from the lower level to learn action models and symbolic options with minimal human knowledge. After that, the meta-controller chooses an option according to the plan generated from the planner with the learned action models and assigns it to the lower level. By interacting with the environments, the lower level learns a policy to reach the assigned option and sends the collected experience to the higher level. 
This cross-fertilization structure not only helps alleviating the sparse and delayed reward problem but also improves the data efficiency. 

Inspired by our previous work on learning hierarchical task networks (HTNs) \cite{DBLP:conf/ijcai/ZhuoHHYM09,DBLP:journals/ai/ZhuoM014,DBLP:conf/aaai/XiaoWZHPC20}, we conjecture that the hierarchical structures of actions can be leveraged to help improve the exploration efficiency and the action model learning accuracy. We thus, in this work, extend our previous work \cite{DBLP:conf/aaai/JinMJZCY22} into learning hierarchical task network models (besides action models), namely {\oursh}, as well as presenting more details about our previous work. With the hierarchical task network models learnt, we can use both task decomposition relations and action models to better guide the exploration of actions in reinforcement learning. We exhibit that our approach significantly outperforms our previous work that only learns action models, when the scale of the environment is large (with respect to the number of objects and their relationships).  

Different from previous hierarchical task network models learning approahces \cite{DBLP:conf/ijcai/ZhuoHHYM09,DBLP:journals/ai/ZhuoM014,DBLP:conf/aaai/XiaoWZHPC20} and action models learning approaches \cite{DBLP:journals/ai/YangWJ07,DBLP:journals/ai/ZhuoYHL10,DBLP:conf/aips/ZhuoYPL11,DBLP:conf/aaai/ZhuoNK13,DBLP:conf/ijcai/ZhuoNK13,DBLP:conf/ijcai/ZhuoK13,DBLP:journals/ai/Zhuo014,DBLP:conf/aaai/Zhuo15,DBLP:journals/ai/ZhuoK17}, which assume the number of task models (a task model is composed of preconditions and effects of the task) and action models to be learnt is known beforehand, both of our {\ours} and {\oursh} do not know exactly ``how many'' and ``what'' task models and action models to be learnt from the environment. We expect the agent continuously creates new hierarchical task models and action models (as well as task decomposition methods) via interactions with the environment, and exploits the new models to guide the exploration of actions to build policies creatively---we consider this as the \emph{creativity} property of an AI agent. We claim that an agent with such creativity can build better policies with respect to transferability, interpretability, and data-efficiency. 

We summarize our contribution as follows:
\begin{itemize}
    \item Our work is the first one to learn action models, hierarchical task network models and option models automatically without being told any knowledge of these models and simultaneously learn RL policies.
    \item We propose a symbolic reinforcement learning framework capable of providing transferability, interpretability, and improved data-efficiency.
    \item The symbolic option learned by {\ours} and {\oursh} is more general, which can correspond to more than one action model.
\end{itemize}

\section{Preliminaries}
In this section we establish relevant notation and briefly introduce key aspects of symbolic planning and reinforcement learning.

\subsection{Symbolic Planning with PDDL}
In PDDL language \cite{DBLP:journals/ai/Gerevini20,PDDL}, states are represented as set of propositions and we call it symbolic states throughout the paper to distinguish \zhma{them} from states in RL. 
Propositions represent the properties of the world and in the \zhma{symbolic} state $s$, proposition $p\in s$ if $p$ is true otherwise $(not \  p)\in s$. An action description called action model is a tuple $(name,pre^+,pre^-,eff^+,eff^-)$, where $name$ is the name of the action,  $(pre^+,pre^-)$ are the preconditions and $(eff^+,eff^-)$ are the effects. As shown in Fig.\ref{fig:actionmodel}, the action model describes that when the agent walks from the middle ladder to the right ladder, the key keeps still and the reward remains unchanged.
If $pre^+ \subset s$ and $s\cap pre^- = \emptyset$, then we can execute action $a$ and obtain the next state 
\[s'=((s - eff^-)\cup eff^+).\]

The planning domain $D=(P,A)$ includes the proposition set $P$ \zhma{and the action set $A$, which} \zhmap{describe the state space and the action space, respectively.}
\zhmap{A tuple $(s,a,s')$ \zhma{describes a} symbolic transition \zhma{from state $s$ to state $s'$ after executing action $a$}. We define a planning problem denoted as a triple $(I,P,A,G)$, of which $I$ is an initial state and $G$ is a goal state. The solution to this problem is called a plan $\pi$, which is a sequence of actions. After executing the plan, we can obtain a symbolic transition trace from $I$ to $G$. To obtain such a plan with the maximum reward, we use a planner called Metric-FF \cite{MetricFF02}, which can handle planning problems with continuous metrics.}



\subsection{Hierarchical Task Networks Planning}

\subsection{Reinforcement Learning}
A Markov Decision Process (MDP) is defined as the tuple $(\widetilde{S},\widetilde{A},P_{\widetilde{s}\widetilde{s'}}^{\widetilde{a}},r^{\widetilde{a}}_{\widetilde{s}},\gamma)$ where $\widetilde{S}$ and $\widetilde{A}$ denote the state space and action space, respectively, $P_{\widetilde{s}\widetilde{s'}}^{\widetilde{a}}$ provides the transition probability of moving from state $\widetilde{s}\in \widetilde{S}$ to state $\widetilde{s'} \in \widetilde{S}$ after taking action $\widetilde{a} \in \widetilde{A}$, $r^{\widetilde{a}}_{\widetilde{s}}$ is the immediate reward obtained after performing action $\widetilde{a}$ \zhmap{at} state $\widetilde{s}$ and $\gamma \in [0,1)$ is a discount factor.
The task of RL is to \zhmap{obtain} a policy \[\pi:\widetilde{S}\rightarrow \widetilde{A}\] that maximizes the expected return \[V_{\pi}(\widetilde{s})={\mathbb E_\pi}[\sum_{t=0}^{\infty}\gamma^{t}r_{t} \mid \widetilde{s}_0=\widetilde{s}]\] 
where $r_t$ is the reward at time step $t$ received by following $\pi$ from state $\widetilde{s}_0=\widetilde{s}$. 
The state-action value function is defined as follows: 
\[Q_{\pi}(\widetilde{s}, \widetilde{a})={\mathbb E_\pi}[\sum_{t=0}^{\infty}\gamma^{t}r_{t} \mid \widetilde{s}_0=\widetilde{s}, \widetilde{a}_0=\widetilde{a}].
\]

\subsection{Option Framework}
Hierarchical Reinforcement Learning (HRL) extends RL with temporally macro actions that represent high-level behaviors. 
The option framework \cite{Sutton1999} models macro actions as options.
In particular, an option $o$ is defined as $(I_o(s),\pi_o(s),\beta_o(s))$, where initiation condition $I_o(s)$ determines whether option $o$ can be executed at state $s$, termination condition $\beta_o(s)$ determines whether option execution terminates at state $s$ and $\pi_o(s)$ is a policy mapping state $s$ to a low-level action. 
In this framework, an agent learns to choose an optimal option to be executed in high level, i.e. the meta controller level, and low level, i.e. the controller level that learns optimal policies to reach the option. An explicit assumption is that the set of options is predefined by human experts.


\section{Problem Formulation}
We define the reinforcement learning environment by a tuple $\langle\mathcal{P}^{sym}, F, \mathcal{M}\rangle$, where $\mathcal{P}^{sym}=\langle I, G, P, A\rangle$ is a symbolic planning problem,  $\mathcal{M}=\langle\widetilde{S}, \widetilde{A}, \widetilde{R}, \widetilde{P}, \gamma\rangle$ is a MDP problem, and $F$ is a function mapping a low-level state $\widetilde{s}$ to high-level symbolic state $s$. In symbolic planning problem $\mathcal{P}^{sym}$, $I$ is an initial state. $G$ is a goal state. $P$ is a set of propositions represented by planning language PDDL with prior knowledge, and it is used to describe symbolic states $S$, where $S \subseteq 2^P$. The initial state $I$ and goal $G$ satisfy $I\in S$ and $G\in S$. $A$ is a set of action models that is used to transit a symbolic state to another: $S\times A \rightarrow S$. Each action model is learned by the meta-controller module through symbolic state pairs. In MDP problem $\mathcal{M}$, $\widetilde{A}$ and $\widetilde{S}$ are sets of low-level actions and low-level states, respectively. $\widetilde{R}$ is the reward function. $\widetilde{P}$ is a set of ???
  
Taking the Figure \ref{fig:statemap} as an example, in the game of the Montezuma's Revenge, the low-level state is the picture of the game scene and the high-level symbolic state is composed of the propositions describing the location of the agent and the existence of the key. 

\section{The SORL Framework with Learnt Action Models}
\begin{figure}[ht]
    \centering
    \includegraphics[width=\textwidth]{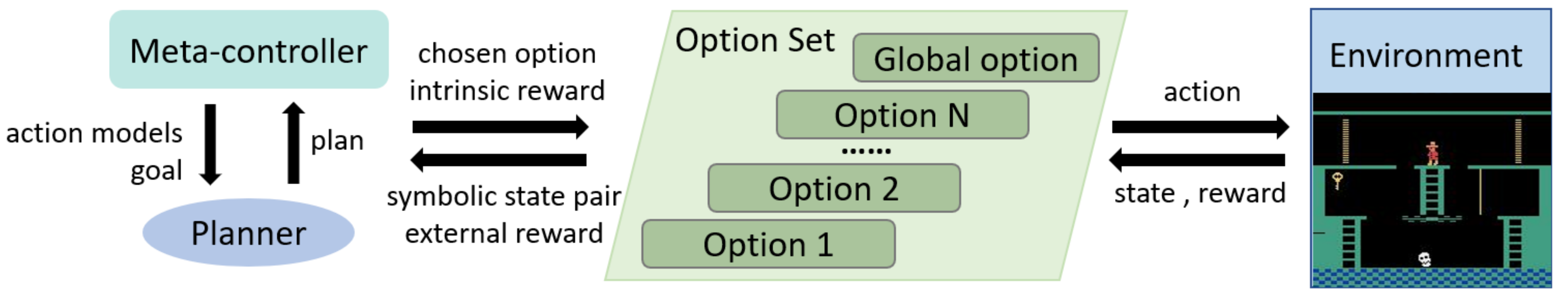}    
    \caption{The SORL framework}
    \label{fig:framework}
\end{figure}

This framework aims to learn action models, which can be utilized by a symbolic planner to generate a sequence of options and achieve the maximal cumulative reward.

As shown in the Figure \ref{fig:framework}, the SORL framework includes three components: (1) a planner for generating plans, (2) a meta-controller for generating action models, goals and choosing the goal option, and (3) an option set for interacting with the environment. The meta-controller first takes the symbolic state pairs and their external rewards as input and outputs action models and a goal. Noted that the state pairs set are empty in the beginning. Then the planner takes the action models as input and computes a plan. Next, the meta-controller receives the plan from the planner and chooses an option. Each option in option set can be regarded as an agent. The chosen agent keeps interacting with the environment until accomplishing the option or reaching the maximal steps, and the low-level state traces will be transformed into symbolic state pairs by the label function $F$ and sent back to meta controller. The meta-controller continues learning action models and symbolic options from gained symbolic state pairs and external rewards. We repeat these procedures $num\_episodes$ times. With the proceeding of learning, our approach keeps updating action models and planning goals and the planner is able to generate plans achieving better rewards.  

\subsection{Option Set}

\paragraph{Symbolic Option}
In this paper, we propose a novel option framework which is called \emph{symbolic option}. A symbolic option is computed by symbolic state pairs gained from trajectories instead of manual setting in advance, 
\zhmap{requiring less prior knowledge in our approach.}
We define a symbolic option by $so = (pre,\pi,eff)$. 
\zhmap{$\pi$ is a low-level policy.}
\zhmap{$pre$ is an union of preconditions, including $pre^+$ and $pre^-$. It's created and updated when the meta-controller generates action models.}
\zhmap{Similarly, $eff$ is composed of $eff^+$ and $eff^-$, describing the effects of the symbolic option.}
As for a symbolic option $so$ and a high-dimension state $\widetilde{s}$, we compute initiation condition $I_{so}(\widetilde{s})$ by Equation (\ref{eq:initiation_condition}) and termination condition $\beta_{so}(\widetilde{s})$ by Equation (\ref{eq:termination_condition}). \zhmap{A} symbolic option \zhmap{can} be executed based on $\widetilde{s}$ only if $I_{so}(\widetilde{s}) = True$. \zhmap{Similarly, it terminates only if  $\beta_{so}(\widetilde{s}) = True$.}
\begin{small}

\begin{equation}
I_{so}(\widetilde{s}) =\left\{
\begin{array}{rcl}
True    &   pre^+ \subset F(\widetilde{s}),F(\widetilde{s})\cap pre^- = \emptyset\\
False   &   otherwise
\end{array} \right.
\label{eq:initiation_condition}
\end{equation}

\begin{equation}
\beta_{so}(\widetilde{s}) =\left\{
\begin{array}{rcl}
True    &   eff^+ \subset F(\widetilde{s}),eff^- \cap F(\widetilde{s}) = \emptyset\\
False   &   otherwise
\end{array} \right.
\label{eq:termination_condition}
\end{equation}
\end{small}

Note that the inherent symbolic propositions of our symbolic option provide better interpretability compared to those approaches based on black-box neural networks.

In terms of the low-level policy $\pi$, it can be learned by interacting with the environments with the intrinsic rewards given by the meta-controller.

\paragraph{Global Option}
At the beginning of our algorithm, the option set contains no symbolic options but a global option \[o_G = (I_G(s),\pi,\beta_G(s)),\]  where $I_G(\widetilde{s})\equiv True$, $\beta_G(\widetilde{s}) = True$ if symbolic state changes and $\pi = random(\widetilde{A})$. We use $random(\widetilde{A})$ to indicate that the global option each time \zhmap{chooses} a random action $\widetilde{a} \in \widetilde{A}$. Intuitively, the global option is available for any state and it keeps random\zhmap{ly} exploring until the symbolic state changes. \zhmap{Hence, in order to discover new action models, }
the meta-controller outputs the global option when the plan is empty or all action models in the plan has been executed.

\paragraph{Symbolic State Pair and External Reward}
\zhma{Given an option $o_j$ under state $\widetilde{s}$, the lower level policy interacts with the environment and output a pair of symbolic states $(s_1,s_2)$ and external reward $r_e$,} denoted by \[(s_1,s_2),r_e = ExecuteOption(\widetilde{s},o_j).\]
\zhmap{I}f the chosen option $o_j$ is not available for $\widetilde{s}$, i.e., $I_j(\widetilde{s})=False$, both of the output pair and the reward are $None$. Otherwise, if the chosen option $o_j$ is able to be executed, we let $s_1 = F(\widetilde{s})$ and the policy $\pi_j$ first choose\zhmap{s} an action $\widetilde{a}$ and we can \zhmap{obtain} the next state $\widetilde{s'}$ and its reward $\widetilde{r}$ by interacting with the environments. Then the controller adds experience $(\widetilde{s},\widetilde{a},\widetilde{s'},\widetilde{r})$ to the $o_j$'s replay buffer. \zhmap{We keep executing action by following the low-level policy and update the states and rewards until}
until $\beta_j(\widetilde{s'})=True$ or reaching the maximum steps, which means the option has been successfully executed \zhmap{or not}. Finally, if \zhmap{the option $o_j$ is} successfully executed, \zhmap{we set the output symbolic state pair as $(s_1,s_2)$ of which $s_2=F(\widetilde{s'})$ and the external reward $r_e$ be the accumulated sum of the environment rewards during interacting.}

\subsection{Meta-controller}
In this section, we introduce our Meta-controller in detail. Meta-controller takes symbolic state pairs and their external rewards as input, and first generates action models and a planning goal and then chooses a an option according to the plan from planner. 

\paragraph{Action Model}
Given symbolic state pairs and their rewards,
meta-controller generates action models by \[A,F_{A,O},O = GenerateActionModels(R,O,sr).\] The function indicates it takes a dictionary $R$,an option set $O$ and the success ratio set $sr$ as inputs, and outputs a generated action set $A$, a mapping function $F_{A,O}$ and the updated option set $O$. Dictionary $R$ includes mappings from a symbolic state pair to its external rewards. $F_{A,O}$ transfers action models to options. The success ratio set $sr$ records the \zhmap{percentage} of action models successful executed each 100 times.

As for a symbolic state pair $(s_1,s_{2})_i\in R$, we can get a corresponding action model 
\[a_i=(name,pre^+,pre^-,eff^+,eff^-).\] Noted that the action model and a symbolic state pair is a one-to-one match. Given a state pair $(s_1,s_{2})_i$, the $name$ of $a_i$ is the index of action models, denoted by $act_i$, and $pre^+ =\{p | p \in s_1\}$, $pre^- = \{p | p\notin s_1 \}$. Next we let $eff^+ = s_{2} - s_{1}$ and $eff^- = s_1 - s_{2}$, where $a - b$ is a set subtraction indicating set $a$ subtracts the intersection of set $a$ and set $b$. In order to generate a plan gaining a maximum reward, we use the metric constant $quality$ to denote the cumulative reward of the plan and add the proposition ``$(increase \quad (quality) \quad \rho_i)$'' into $eff^+$. Finally, we get an action which is called $act_i$, and we define the gain\zhmap{ed} reward of $act_i$ by $\rho_i$. To encourage the planner to \zhmap{generate a} plan \zhmap{including} the exploring action model, the reward $\rho_i$ is composed of mean external reward and exploration reward, computed by Equation (\ref{eq:gain_reward}), where $R[(s_1,s_{2})_i]$ is the external rewards list and $r_E$ is the exploration rewards. The exploration rewards is computed by Equation (\ref{eq:exploration_reward}), where c is a constant and  $sr[i]$ is the success rate of $act_i$, which means exploration reward decreases as success rate increases.


\begin{equation}
\rho_i = mean(R[(s_1,s_{2})_i])+r_{E}
\label{eq:gain_reward}
\end{equation}

\begin{equation}
    r_{E} = \left \{
    \begin{array}{ll}
    c(1-sr[i]) &  act_i\ is\  
being\ explored\\
    0 & otherwise
    \end{array} \right.
\label{eq:exploration_reward}
\end{equation}

If \zhmap{there exists} a symbolic option $o_j=(pre_j,\pi_j,eff_j)$ \zhmap{where} $eff_j=eff$ after we attain an action model,
\zhmap{we update $pre_j^+$ to a union of $pre^+$ and $pre_j^+$, and $pre_j^-$ to a union of $pre^-$ and $pre_j^-$.}
Otherwise, we create a new symbolic option $o_j=(pre,\pi_j,eff)$ and add it to the option set $O$. At last, we set the mapping function $F_{A,O}(act_i)=o_j$. During the exploration, \zhmap{w}e explore each action model sequentially, in other words, we repeat exploring $act_i$ until the success rate of $act_0$ to $act_{i-1}$ is higher than the threshold. 

\paragraph{Planning Goal}
\jinkb{Next Meta-controller outputs a goal to guide planner, aiming at generating a plan with a maximal reward. The goal is  a label function $quality>q$, where $q$ is the cumulative external rewards of the plan gained in the last episode. Intuitively, the function constrains the planner to compute a plan with a largest reward compared with the past plans.}

\paragraph{Chosen Option and Intrinsic Reward}
\jinkb{After the planner \zhmap{generates} a plan $\Pi=(a_1,a_2,\dots,a_n)$, as for each action model $a_i$, the meta-controller selects a symbolic option from option set by $o_j=F_{A,O}(act_i)$, and we can get an series of options $(o_0,o_1,\dots,o_n)$. If all action models in $\Pi$ successfully finish, which indicates the chosen symbolic options are executed sequentially and termination conditions are satisfied \zhmap{, then} the meta-controller would choose \zhmap{the} global option $o_G$ \zhmap{to explore the environment thoroughly.}} For each option $o_i = (pre_i,\pi_i,\beta_i)$, we refer to \cite{DBLP:conf/aaai/LyuYLG19} to design intrinsic rewards:
\begin{equation}
r_i(\widetilde{s}) =\left\{
\begin{array}{rcl}
\phi    &   \beta_i(\widetilde{s})=True\\
r       &   otherwise
\end{array} \right.
\label{eq:intrinsic rewards}
\end{equation}
where $\phi$ is a constant and $r$ is the reward gained from the environments when reach state $\widetilde{s}$.

\begin{algorithm}[tb]
\caption{Planning and Learning algorithm for SORL}
\label{alg:SORL}
\textbf{Input}: proposition set $P$, state mapping function $F$, success ratio threshold $\lambda$
\begin{algorithmic}[1] 
\STATE \textbf{Initialization}: option set $O\leftarrow \{ o_G \}$, action models set $A \leftarrow \emptyset $, symbolic state pairs'  external rewards dictionary $R\leftarrow\emptyset$, action models' success ratio set $sr \leftarrow \emptyset$, plan $\Pi_0 \leftarrow \emptyset$, $q\leftarrow 0$
\FOR{t=1,2, \dots, $num\_episodes$}
\STATE Initialize game, get start state $\widetilde{s_0}$, $I \leftarrow F(\widetilde{s_0})$, $\Pi^* \leftarrow \Pi_{t-1}$
\STATE $A, F_{A,O}, O\leftarrow GenerateActionModels(R,O, sr)$
\STATE $G\leftarrow (quality > q)$
\STATE $\Pi_t \leftarrow metricFF.solve(I,P,A,G)$
\STATE \textbf{if} $\Pi_t = \emptyset$ \textbf{then} $\Pi_t \leftarrow \Pi^*$
\STATE $q\leftarrow 0$
\FOR{ $a_i \in \Pi_t$}
\STATE $o_j\leftarrow F_{A,O}[i]$, obtain current state $\widetilde{s}$
\STATE $(s_1,s_2),r_e\leftarrow ExecuteOption(\widetilde{s},o_j)$
\STATE append $r_e$ into $R[(s_1,s_2)]$, $q \leftarrow q+r_e$
\ENDFOR
\WHILE{env isn't terminal}
\STATE obtain current state $\widetilde{s}$
\STATE $(s_1,s_2),r_e\leftarrow ExecuteOption(\widetilde{s},o_G)$
\IF{$(s_1,s_2)$ not in $R$}
\STATE $R[(s_1,s_2)]\leftarrow list(r_e)$
\ELSE 
\STATE append $r_e$ into $R[(s_1,s_2)]$
\ENDIF
\ENDWHILE
\STATE train options in $O$ and calculate $sr$
\ENDFOR
\end{algorithmic}
\end{algorithm}

\subsection{Planning and Learning}
As shown in Algorithm \ref{alg:SORL}, we firstly initialize an option set $O$ only including $o_G$, an empty action model set $A$, an empty dictionary mapping symbolic state pairs to their external rewards $R$, an empty action models success ratio set $sr$ and an empty plan $\Pi_0$. When an episode $t$ begins, we first get a start state $\widetilde{s_0}$ from environment. We then compute the symbolic initial state $I$ by $F$ and record the best plan $\Pi^*$, which is the plan generated in the last episode. Then meta-controller updates action models $A$, symbolic options set $O$, their mapping function $F_{A,O}$ and \zhmap{the} planning goal $G$. Given current action models $A$ and planning goal $G$, Metric-FF planner \cite{MetricFF02} generates a new plan $\Pi_t$ whose quality is higher than the last plan $\Pi_{t-1}$. If $\Pi_t$ is empty, which indicates Metric-FF couldn't find a solution to solve the problem, we let $\Pi_t = \Pi^*$. 

As for each action model $a_i$ in plan $\Pi_t$, meta-controller chooses a corresponding symbolic option $o_j$ by $F_{A,O}$.  Then the controller interacts with environment by performing Deep Q-Learning, executes the action chosen by $o_j$'s inner policy and stores experience into $o_j$'s replay buffer until $o_j$ terminates. After that, we get $o_j$'s initial symbolic state $s_1$ and a terminate symbolic state $s_2$ and an extrinsic reward $r_e$. In this way, we compute symbolic state pairs and their extrinsic rewards one by one and record these mappings by a dictionary $R$. Finally, quality $q$ of plan $\Pi_t$ is defined as the accumulated sum of extrinsic rewards. 

If the environment is not finished after executing $\Pi_t$, the
meta-controller chooses the global option $o_G$ to explore new symbolic states pairs in the environment.
$o_G$ stops exploring when the computed symbolic state changes and we calculate a symbolic state pair $(s_1,s_2)$ and its external reward $r_e$. If $(s_1,s_2)$ is a new symbolic state pair, we add it into $R$. This process repeats until the environment is \zhmap{terminated}. Finally, when an episode ends, we train options in $O$ and calculate success ratio for each action model.

\section{The {\oursh} Framework with Learnt HTN Models}
\begin{figure}[!ht]
    \centering
    \includegraphics[width=0.9\textwidth]{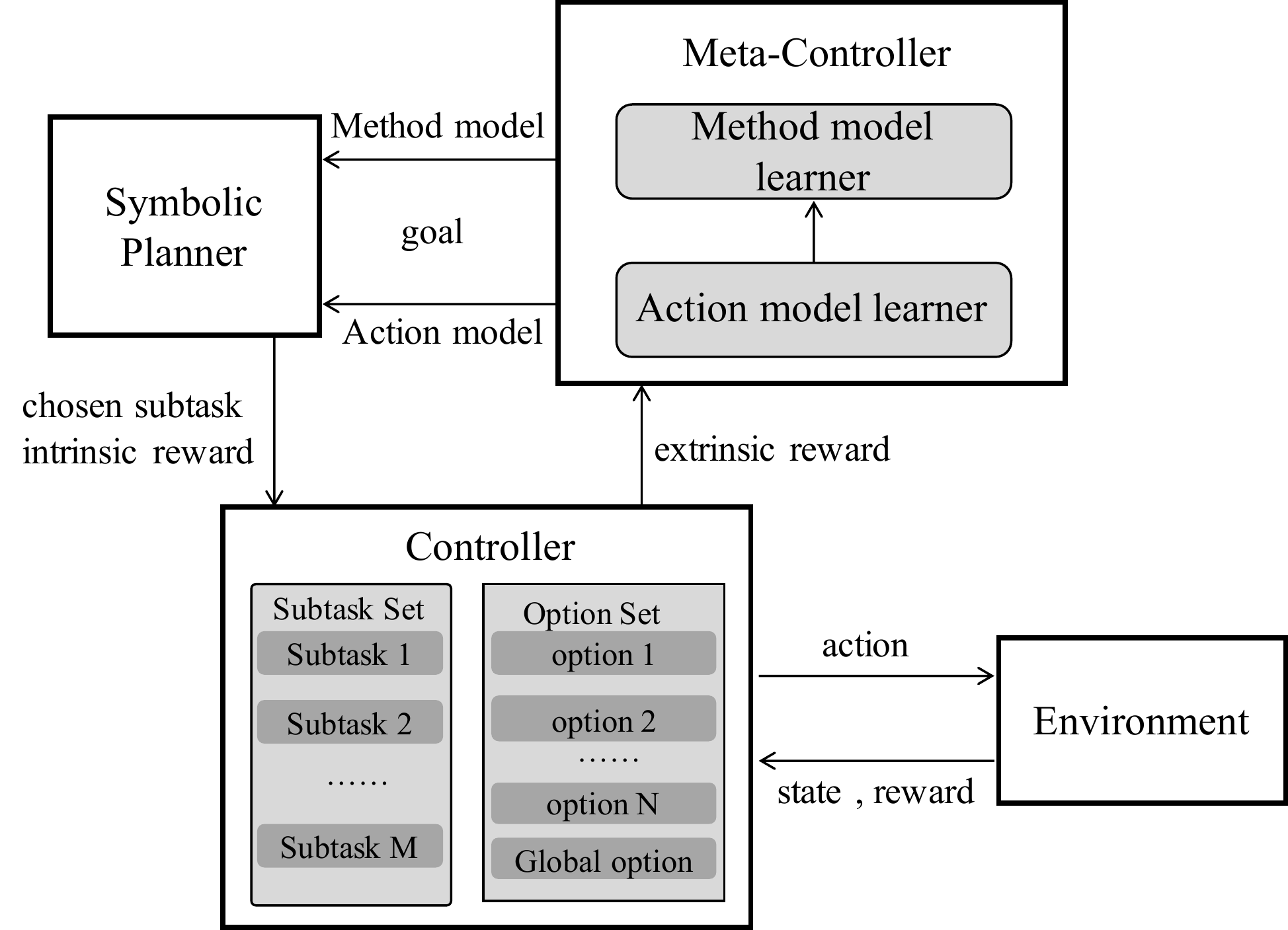}
    \caption{The framework of learning HTN and action models}
    \label{fig:htn}
\end{figure}
To further leverage the task structures for guiding the exploration of actions in reinforcement learning, we build a novel algorithm framework to learn hierarchical task networks based on the action models learnt by Algorithm \ref{alg:SORL}. An framework of learning HTN and action models is shown in Figure \ref{fig:htn}, where the controller module uses the set of (learnt) option models (denoted by "Option Set") to complete the set of subtasks (denoted by "Subtask Set") with interaction to the environment, i.e., executing a low-level "action" and receiving "reward" and new "state" from the environment. After that, the Meta-Controller module receives the extrinsic reward computed by the Controller module and learn hierarchical task network models (i.e., decomposition methods, which is denoted by "Method model learner") and action models (denoted by "Action model learner") based on the extrinsic reward. Finally, the Symbolic Planner module receives the "Method models" and "Action models" from the Meta-Controller module and computes subtasks and intrinsic rewards for the Controller module, with respect to reaching the input goal. In the following, we will address the details of the three modules: Controller, Meta-Controller, and Symbolic Planner. 

Specifically, we provide an algorithm framework, as shown in Algorithm \ref{alg:htn}, to detail the learning procedure of Figure \ref{fig:htn}. In Algorithm \ref{alg:htn}, Steps 4 and 5 aim to learn action models and decomposition methods, i.e., the Meta-Controller module. Step 6 aims to compute a list of subtasks and the intrinsic reward, i.e., the Symbolic Planner module. Steps 9 to 32 aim to complete subtasks with interaction to the environment, i.e., the Controller module. 
\begin{algorithm}[!ht]
\caption{An overview of our HTN-based reinforcement learning}
\label{alg:htn}
\textbf{Input:} tasks $T$, propositions $P$, state mapping function $F$, success threshold $\lambda$\\
\textbf{Output:} action models $A$, method models $M$, options $O$
\begin{algorithmic}[1]
    \STATE \textbf{Initialization:}  $O\leftarrow \{ o_G \}$,  $A \leftarrow \emptyset$,  $M \leftarrow \emptyset$, symbolic state pairs $SP \leftarrow \emptyset$, plan $\Pi_0 \leftarrow \emptyset$, exploration trace $\tau \leftarrow \emptyset$, success ratios of action models $sr \leftarrow \emptyset$
    \FOR{$t=1, 2, \dots, num\_episodes$}
        \STATE Initialize game, get initial state $\widetilde{s_0}$, $I \leftarrow F(\widetilde{s_0})$
        \STATE $A, F_{A,O}, O\leftarrow GenerateActionModels(SP,O, sr)$
        \STATE $M, F_{M,T} \leftarrow GenerateMethodModels(\tau,T, A)$
        \STATE $\Pi_t \leftarrow HTNsolver(I,P,A,M,T)$
        \STATE \textbf{if} $\Pi_t = \emptyset$, let $\Pi_t \leftarrow \Pi_{t-1}$
        \STATE $\tau \leftarrow \emptyset$
    \FOR{ $m_i \in \Pi_t$}
    \STATE $t_j\leftarrow F_{M,T}[m_i]$
    \FOR{ $a_k \in \Pi_t(m_i)$}
    \STATE $\tau \leftarrow \tau + a_k$
    \STATE $o_l\leftarrow F_{A,O}[a_k]$, obtain current state $\widetilde{s}$
    \STATE $(s_1,s_2),r_e\leftarrow ExecuteOption(\widetilde{s},o_l)$
    \STATE If $\beta(t_j) = \TRUE$, let $r_e \leftarrow r_e + R_T$
    \STATE append $r_e$ into $SP[(s_1,s_2)]$
    \ENDFOR
    \ENDFOR
    \WHILE{env isn't terminal}
    \STATE obtain current state $\widetilde{s}$, $s \leftarrow F(\widetilde{s})$
    \STATE $A_s \leftarrow  \{$ operator instances applicable to $s\} $
    \IF{$A \neq \emptyset$}
    \STATE nondeterministically choose an $a_s \in A_s$
    \STATE $\tau \leftarrow \tau + a_s$, $o_m\leftarrow F_{A,O}[a_s]$
    \STATE $(s_1,s_2),r_e\leftarrow ExecuteOption(\widetilde{s},o_m)$
    \STATE append $r_e$ into $SP[(s_1,s_2)]$
    \ELSE
    \STATE $(s_1,s_2),r_e\leftarrow ExecuteOption(\widetilde{s},o_G)$
    \STATE If $(s_1,s_2)$ not in $SP$, let $SP[(s_1,s_2)]\leftarrow list(r_e)$, otherwise, append $r_e$ into $SP[(s_1,s_2)]$
    \ENDIF
    \ENDWHILE
    \STATE train options in $O$ and calculate $sr$
    
    \ENDFOR
\end{algorithmic}
\end{algorithm}

\section{Experiment}
In this section, we evaluate our approach on two domains, Office World and Montezuma's Revenge in terms of data-efficiency, interpretability and transferability. 
\begin{figure}[!ht]
    \centering
    \subfigure[Office World]{
        \label{fig:exp1-1}
        \includegraphics[width = 5cm]{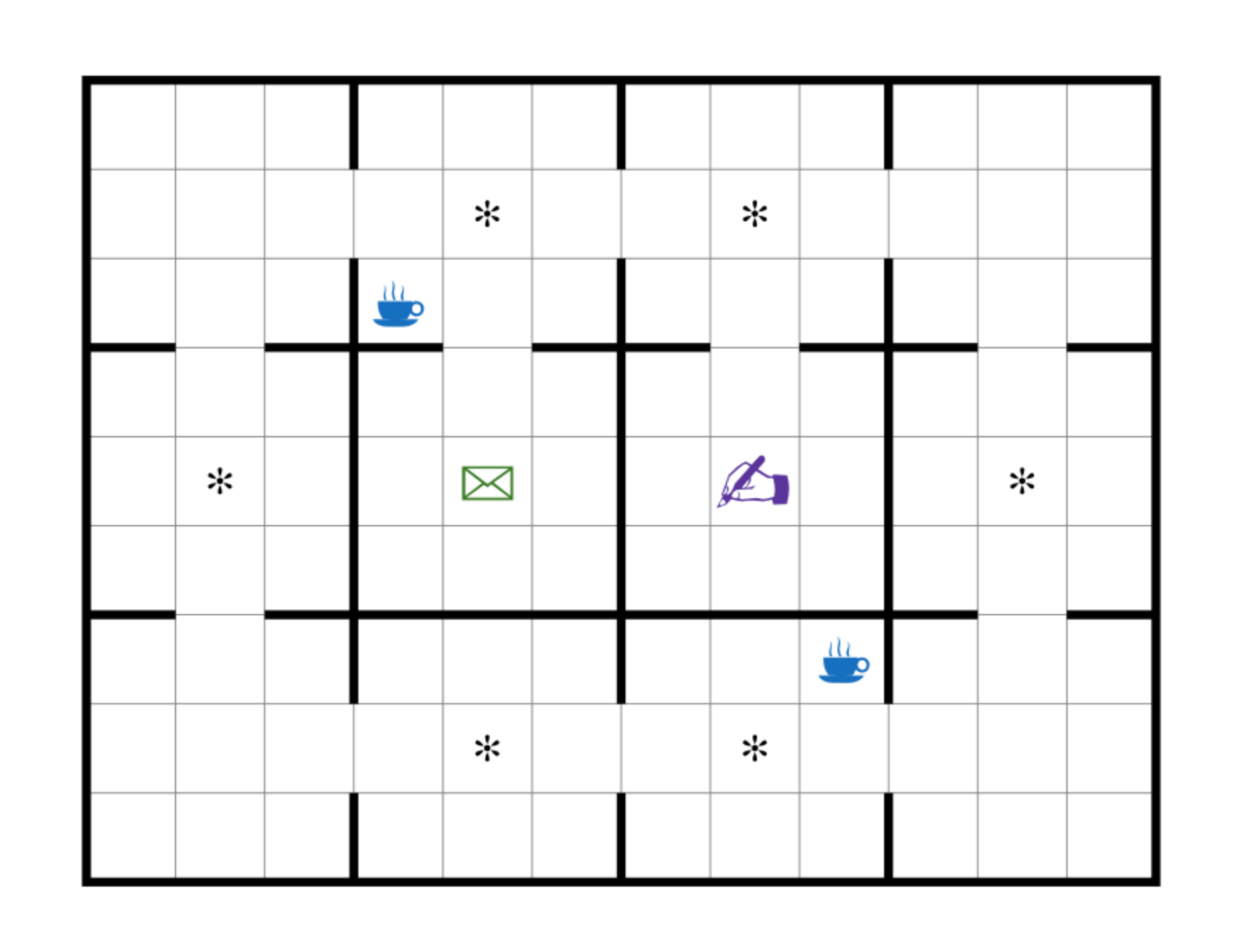}}
    \subfigure[Learning Curve]{
        \label{fig:exp1-2}
        \includegraphics[width = 6.5cm]{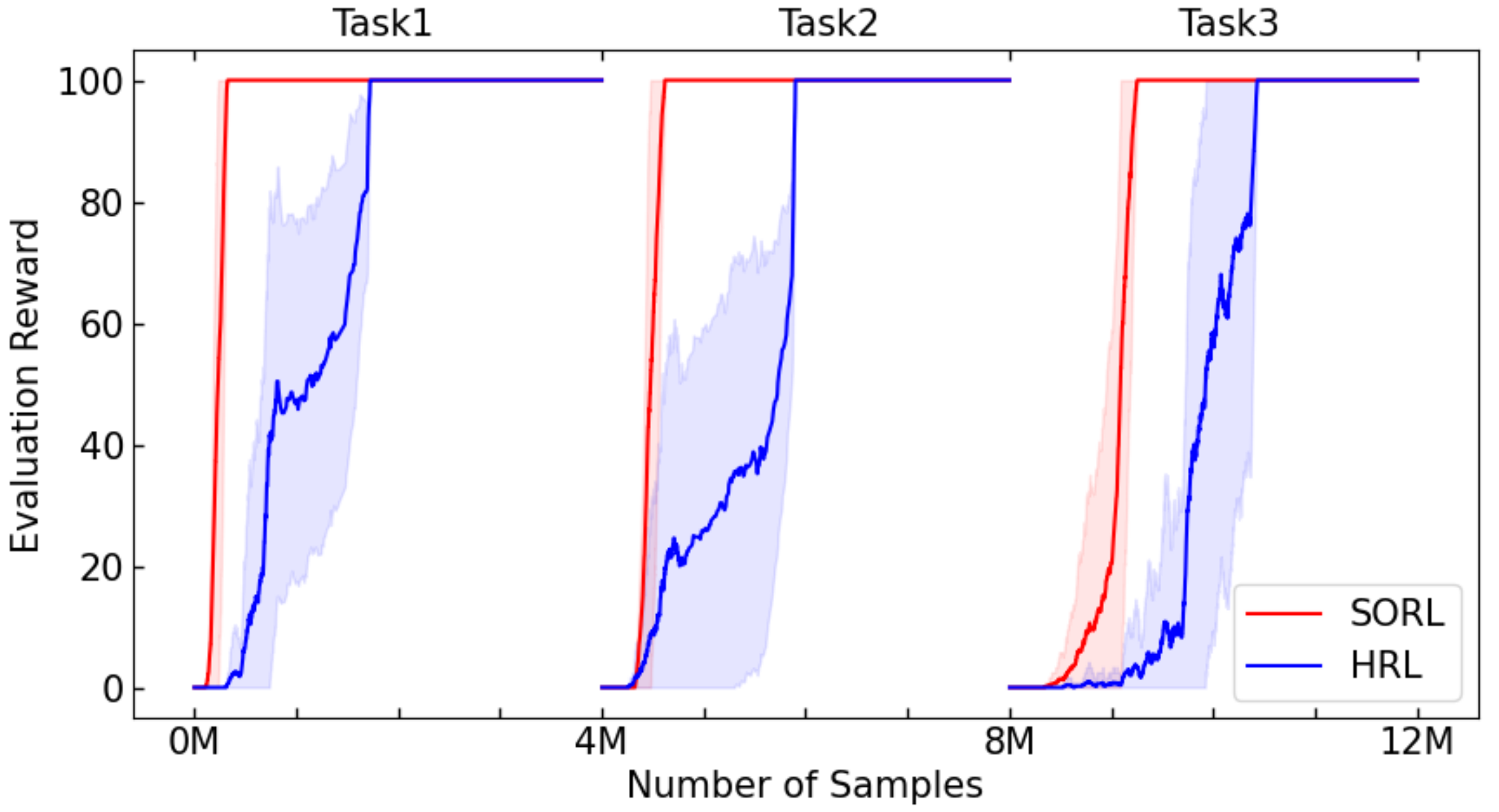}}
    \subfigure[Propositions]{
        \label{fig:exp1-3}
        \includegraphics[width = 5.5cm]{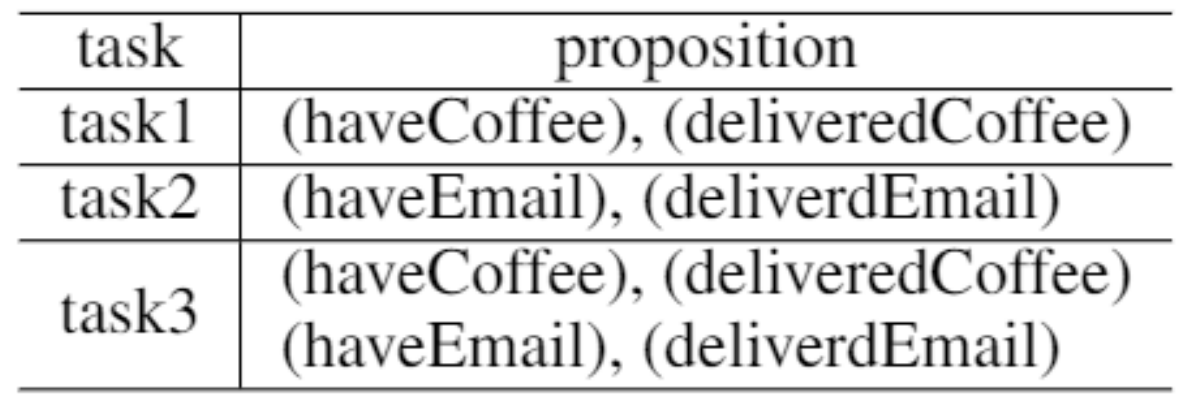}}
    \subfigure[Learned Action Models and Options]{
        \label{fig:exp1-4}
        \includegraphics[width = 11cm]{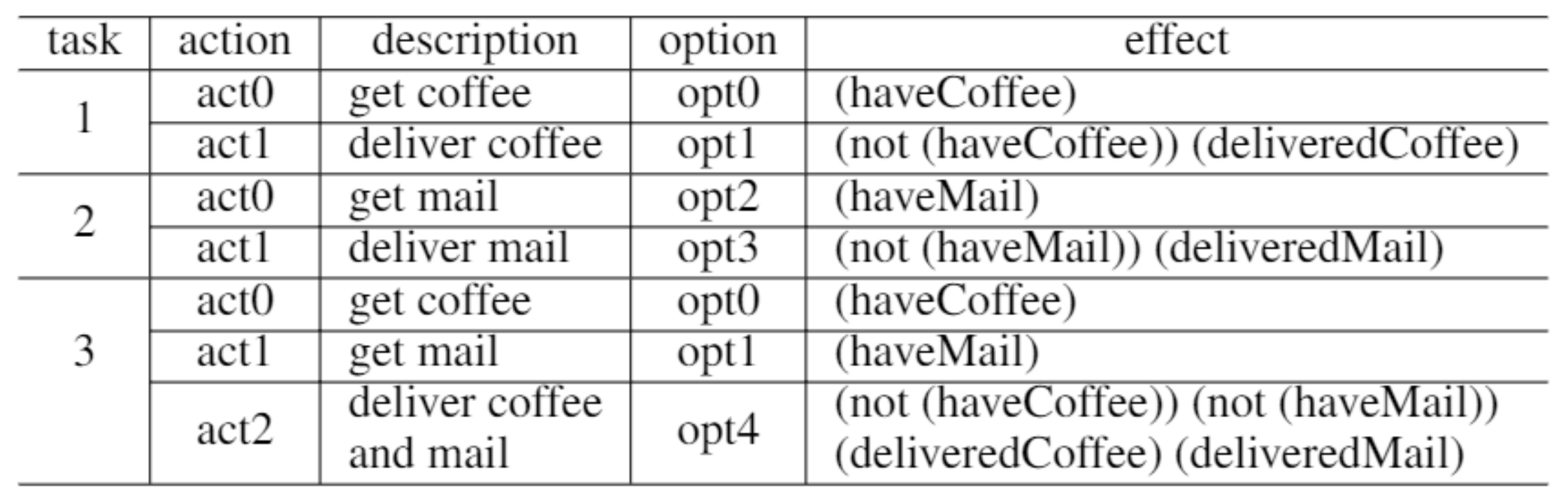}}
    \subfigure[Learning Curve]{
        \label{fig:exp1-6}
        \includegraphics[width =0.5\textwidth]{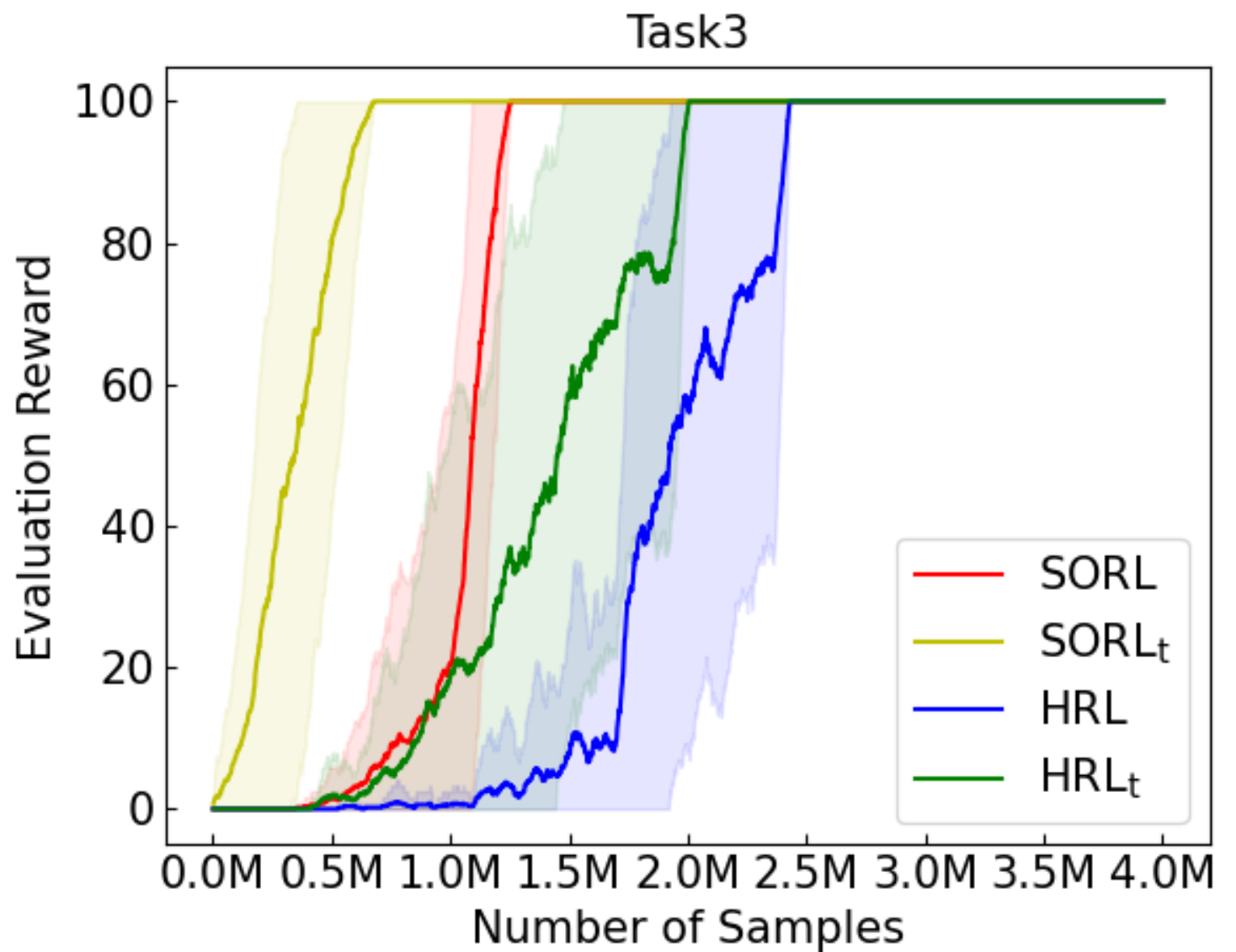}}
    \caption{Experimental Results in the Office World}
    \label{fig:exp1}
\end{figure}

\subsection{Office World}
We first evaluate our approach on \zhma{the} Office World \cite{Icarte2018} which is a simple multitask environment. In this environment, \zhma{being initialized at a random location}, the agent can move \zhma{towards one} of the four cardinal directions. \zhma{Actions are valid}
only if the movement does not go through a wall.
The agent can pick up \zhma{cups of} coffee or mail\zhma{s} when it reaches the cell marked with blue cups or green envelops, respectively. \zhma{He} can deliver coffee or mail to the office \zhma{by reaching the cell marked with a purple hand. The symbol $*$ means the place where the agent can not stay or reach.}

\subsubsection{Setup}
In this environment, the start location of the agent is randomly initialized at every episode. The agent is required to finish three tasks. The first and the second are to deliver a cup of coffee or a piece of mail to the office while the third is to hand both objects to the office. We compared SORL to h-DQN, a goal based HRL \hankz{approach} \cite{DBLP:conf/nips/KulkarniNST16}. Since the state and action space are finite, we choose to implement these two approaches with q-table in both high and low levels.

\subsubsection{Results}
We evaluate our approaches in terms of data-efficiency, interpretability and transferability.

\begin{itemize}
\item \textbf{Data-efficiency} \zhma{In order to validate the data-efficiency, we train these two \hankz{approaches} in the three tasks and compare the corresponding performance at the same interaction steps. To demonstrate the transferability, we train the agent in task 3 along with the options learned in tasks 1 and 2.}
To implement our approach, we design the propositions as \hankz{shown in Fig.} \ref{fig:exp1-3}. As shown in Fig.\ref{fig:exp1-2} , from Task1 to Task3, SORL can get rewards faster than HRL. 

\item{\textbf{Interpretability}}
Fig.\ref{fig:exp1-4} shows action models and symbolic options learned in each task. 
Those action models describe the reason of making decisions at each step in a human understandable way. 
For example, we can explicitly know $act1$ in task1 can be executed when the agent gets coffee and does not deliver it to the office, and the agent would deliver the coffee to the office and get a reward of 100 when $act1$ is executed.

\item{\textbf{\hankz{Transferability}}} By utilizing the options learned in tasks 1 and 2, we test the tranferability of SORL and H-DQN in Task 3 and denote them as $SORL_t$ and $HRL_t$. As shown in Fig\ref{fig:exp1-3}, the performance of SORL and HRL is improved when transferring the learned knowledge. It verifies that compared to SORL, the converging speed of $SORL_t$ improves dramatically with only half of samples. We conjecture that the SORL is able to transfer the learned knowledge into other unseen environments.
\end{itemize}

\subsection{Montezuma's Revenge}
\begin{figure}[!ht]
    \centering
    \subfigure[Learning Curve]{
        \label{fig:exp2-1}
        \includegraphics[width = 0.5\textwidth]{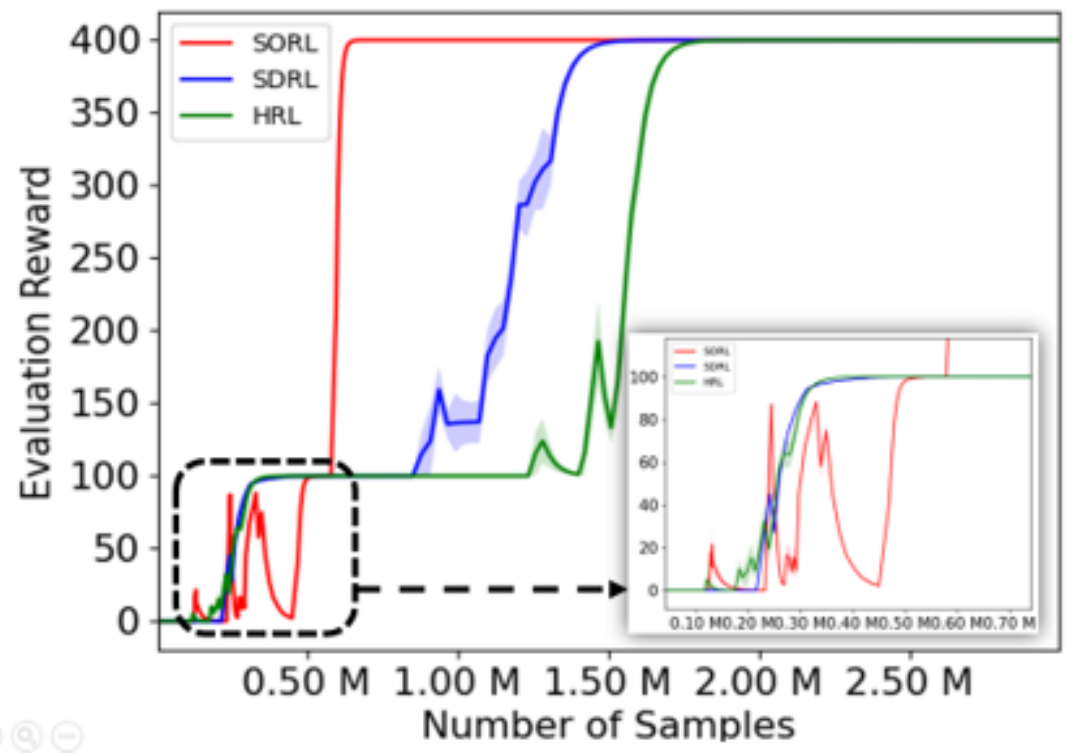}}
    \subfigure[Learned Action Models]{
        \label{fig:exp2-4}
        \includegraphics[width = 11cm]{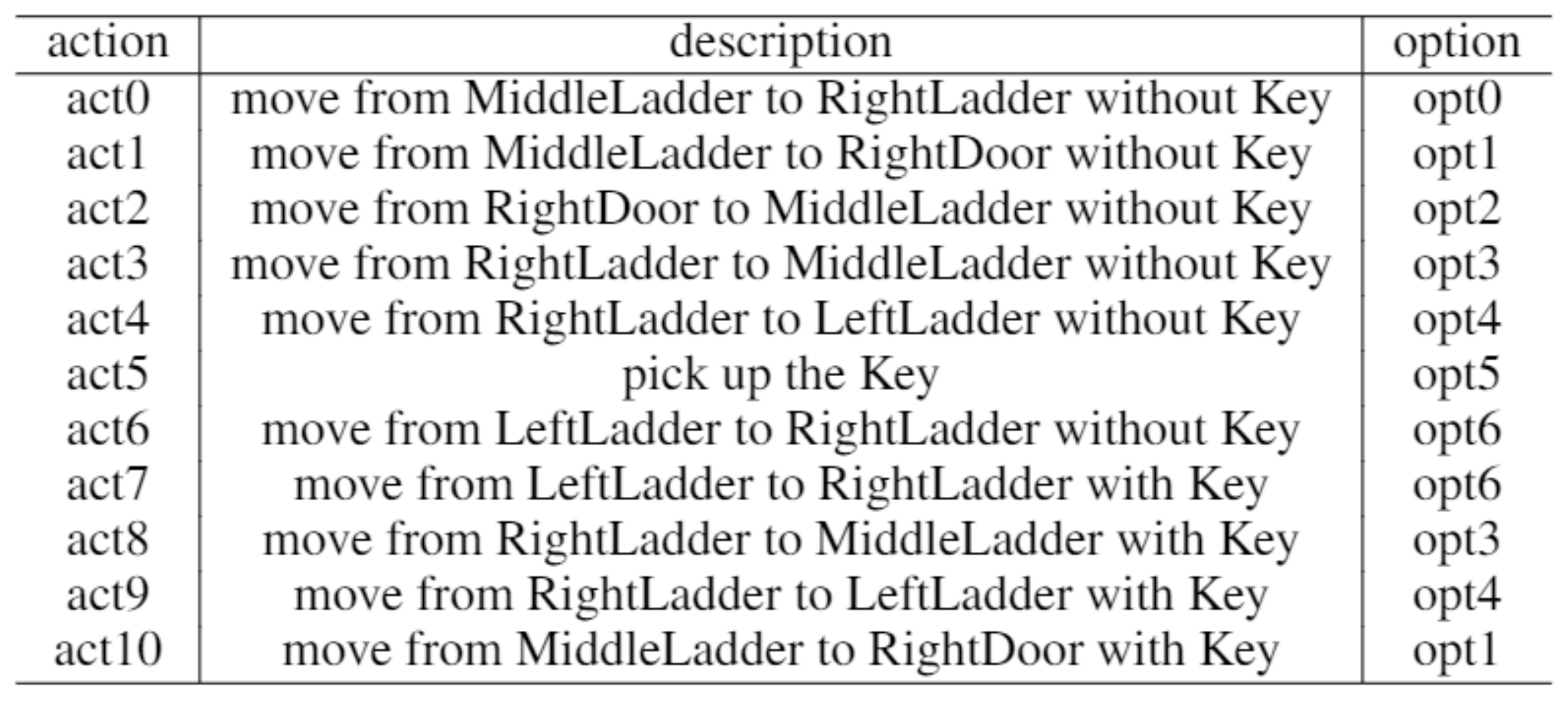}}
    \caption{Experimental Results in Montezuma's Revenge}
    \label{fig:exp2}
\end{figure}

Montezuma's Revenge is an Atari game with sparse \zhma{and} delayed rewards. It requires the player to navigate through several rooms while collecting treasures. We conduct our experiments based on the first room shown in Fig.\ref{fig:exp2-1}. 
In this room, the player only obtains positive rewards when it fetches the key (+100) or opens a door (+300). Otherwise, the player would not receive any reward signal. The optimal solution is to climb down the ladders to obtain the key, then return back to the platform and open a door, resulting in a maximum reward (+400).
\subsubsection{Setup}
We compare our approach with HRL \cite{DBLP:conf/nips/KulkarniNST16} and SDRL \cite{ DBLP:conf/aaai/LyuYLG19} as baselines, where SDRL is an approach that combines symbolic planning and RL with excellent results in complex environments with sparse rewards. 
SORL can automatically learn the action models while they are pre-defined by experts in SDRL. Besides, the option model can correspond to multiple action models in SORL while one in SDRL.
We implement these approaches under 
an option-based HRL framework.
In terms of the low level, we follow the network architecture used in
 \cite{DBLP:conf/nips/KulkarniNST16} and train this network with double-Q learning \cite{DDQN2016} and prioritized experience replay \cite{Prioritizedexperience}. 
Besides, both SORL and SDRL use a planner to generate high level policy while HRL utilizes a neural network. The intrinsic reward follows \ref{eq:intrinsic rewards} with $\phi = 100$. The maximum steps in an episode and the threshold of success rate are set to be 500 and 0.95, respectively. To describe the environment, we abstract four local propositions (e.g., MiddleLadder, RightDoor, LeftLadder and RightLadder) and an object (Key).

\subsubsection{Results}
\begin{figure}[!ht]
    \centering
    \includegraphics[width = 0.6\textwidth]{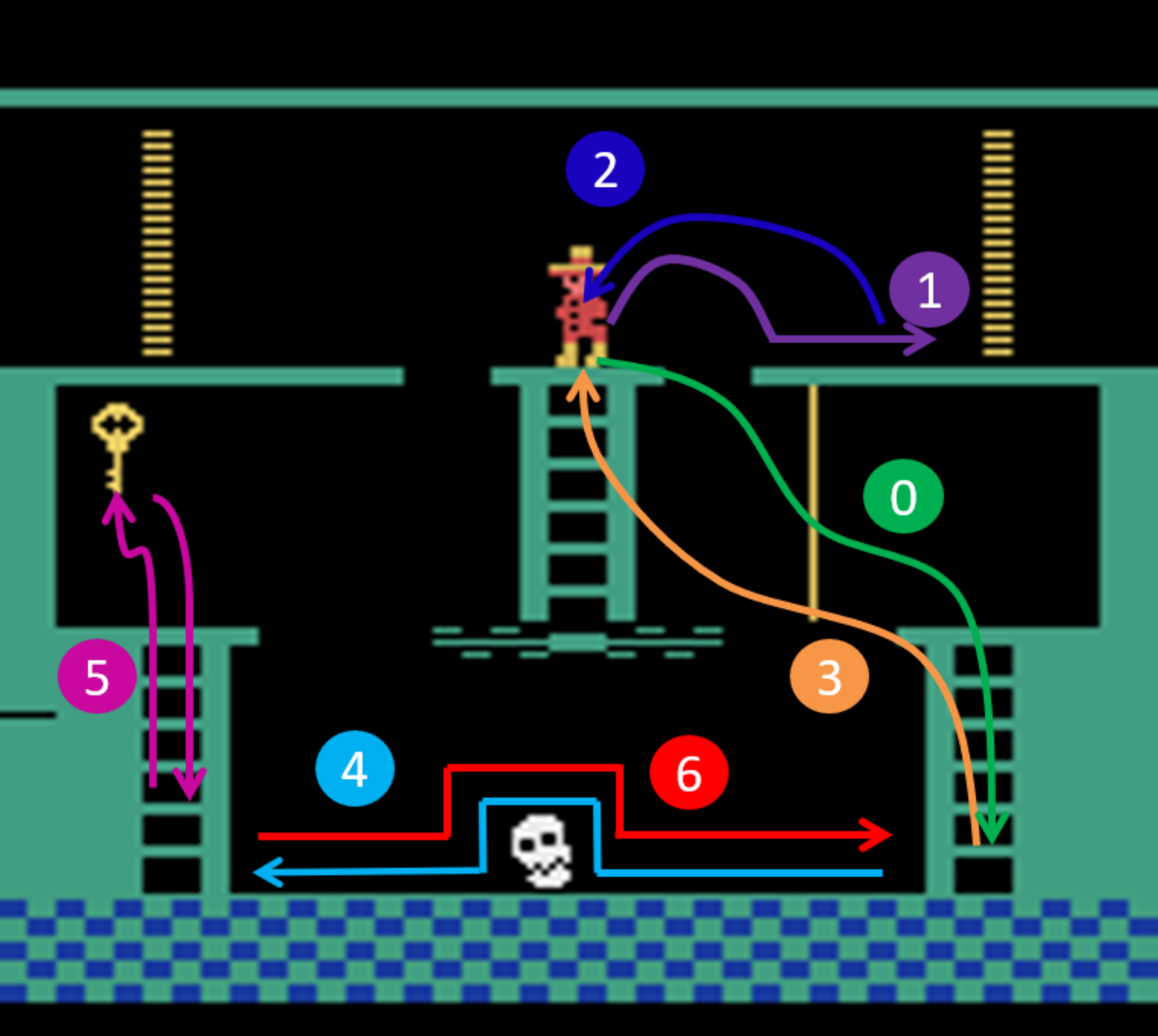}
    \caption{learned Symbolic Options}
    \label{fig:exp2-3}
\end{figure}

We present the experimental result\zhma{s} in Fig.\ref{fig:exp2}. 
It is evident that SORL can achieve the maximum reward (+400) in 0.7M samples while both SDRL and HRL need more than 1.5M samples, indicating the superior data-efficiency of SORL.
However, to pick up the key (reward +100), SORL needs to interact with the environment with more than 0.3M steps, at which SDRL and HRL fall into the local optimum. This is because SORL randomly explores symbolic options and it is easier to find options closer to the starting point.
After finding these options, SORL would train them sequentially instead of directly learning options on the path of getting the key.
One option model corresponds to one action model in SDRL while several action models in SORL. The ability of reusing the learned symbolic options enables SORL to converge faster than SDRL.
Take Fig.\ref{fig:exp2-3} as an example, the $opt1$ representing the move from middle ladder to right door, is firstly trained at the beginning when the player does not get the key. After the player picks up the key, SORL only needs a small amount of data to fine-tune $opt1$ when the player moves from the middle ladder to right door with a key. However, both SDRL and HRL start training the options after the player moves to the middle ladder with a key, consuming more interaction resources.
Different from option-based HRL and SDRL, 
SORL 
can learn the initial and termination condition of symbolic options automatically.
\zhma{The action models used in SDRL need to be constructed by human in advance while they are learned from the trajectories in SORL, saving labour resources.}

\begin{figure}[!ht]
    \centering
    \includegraphics[width = 0.8\textwidth]{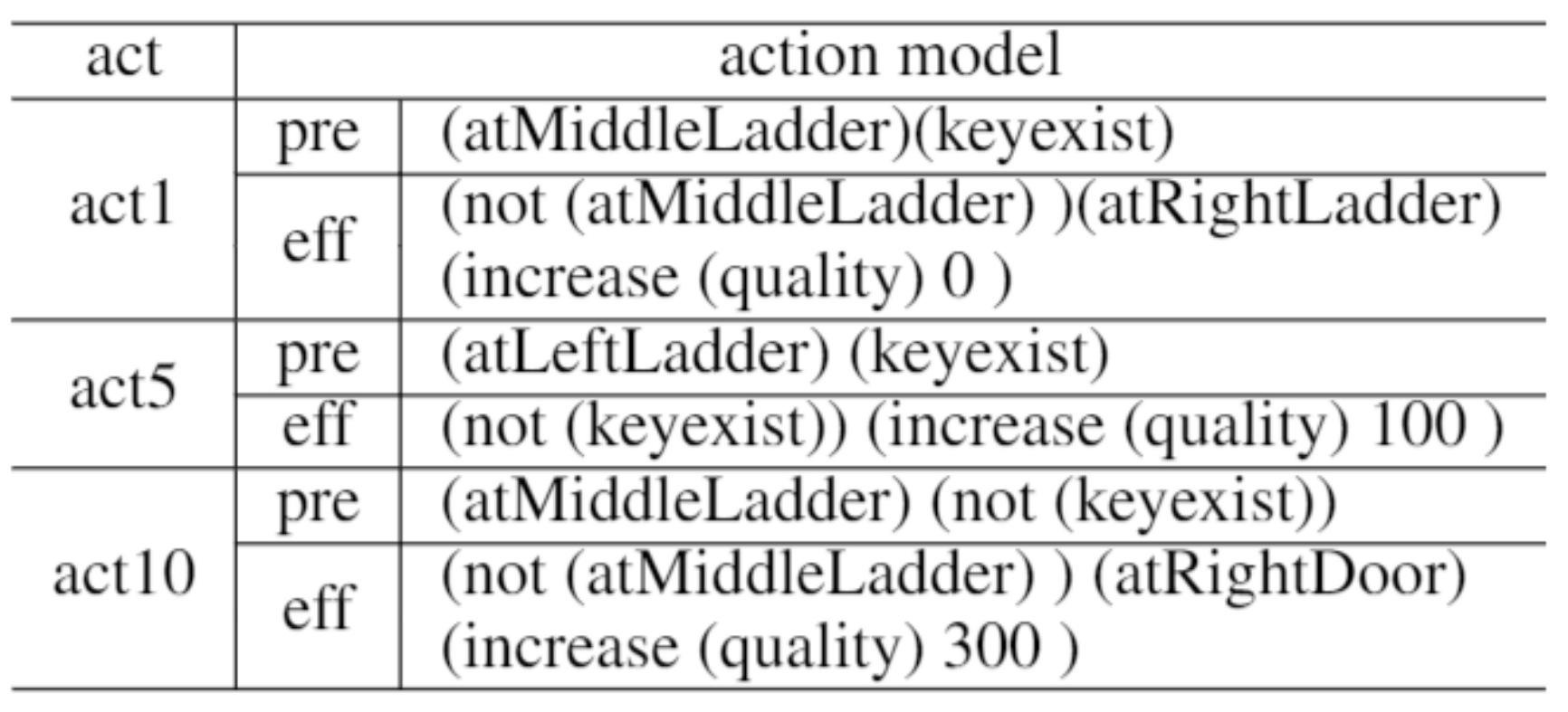}
    \caption{Examples of Learned Action Models}
    \label{fig:exp2-2}
\end{figure}

We present some of the learned action models in Fig.\ref{fig:exp2-2} and the effects of symbolic options in Fig.\ref{fig:exp2-3}.
\zhma{Fig. \ref{fig:exp2-2} describes the preconditions and effects of each action model and we can see that if the player is at LeftLadder and the key exists, then the player can obtain a key and  reward (+100) by executing action5. Fig. \ref{fig:exp2-3} shows the learned options in SORL and the order of options actually does not match the optimal order because SORL randomly explore the environment and options 0-3 are easier to learn.}
\zhma{We describe the meaning} of all learned action models and their corresponding options in Fig. \ref{fig:exp2-4}. \zhma{It is easy to see that act7 to act10 correspond to the options explored before, so these options can be reused to improve the data-efficiency.}

\section{Conclusions} 
In this paper, we propose two novel frameworks {\ours} and {\oursh} \zhma{which} can automatically learn action models, hierarchical task models, and symbolic options from the trajectories and the symbolic planner can instruct RL to explore efficiently in \zhma{environments with sparse and delayed rewards.} 
\zhma{Compared with other approaches, the experimental results demonstrate the better sampling efficiency of our approach. Moreover, {\ours} and {\oursh} require less prior knowledge and provides interpretability and transferability by the learned action models and symbolic options.} 
In the future, it would be interesting to investigate possibility of learning more expressive planning models in real-world applications, as well as different learning mechanisms, such as transfer learning  \cite{DBLP:journals/ai/Zhuo014,DBLP:conf/aaai/ShenZXZP20}.

\section*{Acknowledgement}
This research was funded by the National Natural Science Foundation of China (Grant No. 62076263), Guangdong Natural Science Funds for Distinguished Young Scholar (Grant No. 2017A030306028), Guangdong special branch plans young talent with scientific and technological innovation (Grant No. 2017TQ04X866), Pearl River Science and Technology New Star of Guangzhou and Guangdong Province Key Laboratory of Big Data Analysis and Processing.

\bibliographystyle{elsarticle-harv}
\bibliography{ref,aij}

\end{document}